\documentclass{ifacconf}

\usepackage{graphicx}
\usepackage{natbib}
\usepackage{caption}
\usepackage{subcaption}
\usepackage{makecell}
\usepackage{booktabs}
\usepackage{pifont}
\usepackage{xcolor}
\usepackage{url}
\usepackage{placeins}

\begin{document}
\begin{frontmatter}



\title{Reinforcement Learning in an Adaptable Chess Environment
for Detecting Human-understandable Concepts} 

\author[First]{Patrik Hammersborg$^1$} 
\author[First]{Inga Str{\"u}mke$^2$} 

\address[First]{Department of Computer Science,
    Norwegian University of Science and Technology,
    Trondheim, Norway \\
    $^1$patrih@stud.ntnu.no \quad
    $^2$inga.strumke@ntnu.no
    }
This work has been submitted to IFAC for possible publication.

\begin{abstract}
Self-trained autonomous agents developed using machine learning are showing great promise in a variety of control settings, perhaps most remarkably in applications involving autonomous vehicles. The main challenge associated with self-learned agents in the form of deep neural networks, is their black-box nature: it is impossible for humans to \emph{interpret} deep neural networks. Therefore, humans cannot directly interpret the actions of deep neural network based agents, or foresee their robustness in different scenarios. In this work, we demonstrate a method for probing which \emph{concepts} self-learning agents internalise in the course of their training. For demonstration, we use a chess playing agent in a fast and light environment developed specifically to be suitable for research groups without access to enormous computational resources or machine learning models.
We provide code for the environment and for self-play learning chess agents, as well as for the state-of-the-art explainable AI (XAI) method of concept detection. We present results for different chess board sizes, discussing the concepts learned by our chess agents in light of human domain knowledge. 
\end{abstract}


\end{frontmatter}

\section{Introduction}
Autonomous agents are increasingly developed using machine learning (ML), most often with a reinforcement learning (RL) approach. In the RL setting, ML based agents explore their environment freely, with the aim of performing a task or solving a problem. Many control problems greatly benefit from the RL paradigm, since it alleviates many of their inherent challenges. It lets the agent itself be in charge of data collection, meaning that problems with large state spaces can be sufficiently explored without having to hand-tailor a data set to the problem. This can make the agent more robust, despite the fact that most real-life control environments cannot be fully modelled. RL methods also have the benefit of a goal-oriented approach to problems, meaning that we do not need to specify individual decisions in order to reach the end-goal. 

While RL for fitting neural network models is an effective approach for creating autonomous agents, it introduces a problem related to human understanding: the resulting models are not interpretable to humans, opening up questions regarding \emph{what} the agent has learned, and whether it has internalised knowledge human domain experts know to be important. Based on the data processing inequality (see e.g.~\cite{dataprocessinginequality}), we know that an ML model cannot add to the information contained in its input data.  An important goal for explainable AI (XAI) in the context of autonomous agents is therefore to detect what knowledge these agents have conceptualised. We argue that investigating what a model has learned and whether this aligns with human domain knowledge is crucial for human oversight, and ensuring the stability and safety of of autonomous agents.

The game of chess is the most thoroughly explored environment in the history of artificial intelligence (AI).
We argue that chess is also highly suitable for developing and testing both training and explanation methods for control, for the following reasons. 
Firstly, the dimensionality of the game is too large for complete exploration of the state space. The RL agent must therefore find an exploration exploitation trade-off in order to succeed in solving its task, as is the case in most real-life control settings. 
Secondly, the game is extremely complex, meaning that the objective shifts from finding an optimal solution, to producing efficient heuristics for approximating optimal play. 
Thirdly, the entire game with all its complexity can be completely simulated in a fast and efficient manner, and -- with the environment provided in this work -- with the desired dimensionality and thus computational cost. This means that it provides a problem that is complex enough for these methods to be relevant, while not being exceedingly computationally expensive, like many other control-problems.
Finally, the game features a large amount of available expert knowledge, meaning that explanations of the agent can be readily evaluated.  

Today, AlphaZero \citep{alphazero_original} is the state-of-the-art ML based chess playing agent, demonstrating RL agents to be capable of outperforming human experts, and matching the strength of the best chess engines in existence. Recent work, available in the form of a preprint \citep{alphazero_concepts_2021}, demonstrates that it is possible to probe which \emph{human concepts} AlphaZero has learned through self-play. This methodology is likely to prove itself highly relevant in control contexts, providing us with the opportunity to investigate what domain knowledge black-box autonomous agents have acquired.

A considerable challenge in this regard is that the re-implementation of the kind of training loop required for such investigation is beyond the computational resources available to most research groups. 
The classical game of chess is 8x8 dimensional, and requires the evaluation of approximately \({10^{123}}\) positions in order to determine the value of the initial position \citep{statespacechess}.
It is estimated that AlphaZero performed 44 million self-play games for its peak playing strength. With a large network, this is extremely resource intensive. Additionally, the concept detection method used in \cite{alphazero_concepts_2021} is based on outputs from the intermediary layer blocks of their network, meaning that the dimensionality of the data set depends of the size of the network. Given their specifications of 256x3x3 dimensions per layer, with a specified maximum sample size of the data set at \(5 \cdot {10^5}\), this means one needs to find a least-fit regressor on 2304 variables over such a data set for each such layer.

In order to mitigate this challenge and make this promising venue of investigating XAI methods for detecting concepts learned by RL agents in complex environments, we provide
\begin{itemize}
    \item[\ding{51}] a fast, open-source chess environment for easy simulation of any chess board size, starting position and piece types\footnote{Available piece types are limited to those included in standard 8x8 chess.},
    \item[\ding{51}] fully trained RL agents mastering game-play on smaller chess boards,
    \item[\ding{51}] code for probing concepts learned by trained agents, together with results showing the detected concepts.
\end{itemize}

The paper is organised as follows.
In Sec.~\ref{sec:methods}, we briefly outline the training of the RL agents, present the environment, and describe the concept detection method.
In Sec.~\ref{sec:results}, we present results for training agents in two different environments, and the evolution of the learned concepts throughout their training processes.
In Sec.~\ref{sec:analysis}, we analyse the learned concepts, and how they relate to the agents' learning processes.
In Sec.~\ref{sec:discussion}, we discuss limitations and possible developments of the concept detection methods, including potential relevance to other areas of application.
\section{Method}\label{sec:methods}
\subsection{Training the Agents}\label{sec:agents}
The goal is for an agent to master the game of chess via self-play using reinforcement learning. Our agent is a deep neural network with randomly initialised weights, meaning that it contains no knowledge at the beginning of the training loop. Its ability to play chess arises solely from experience during self-play. Through the exploration of different moves, it learns different playing styles, to recognise useful patterns on the board, and to evaluate board positions.

\begin{figure}
    \centering
    \includegraphics[width=0.35\textwidth]{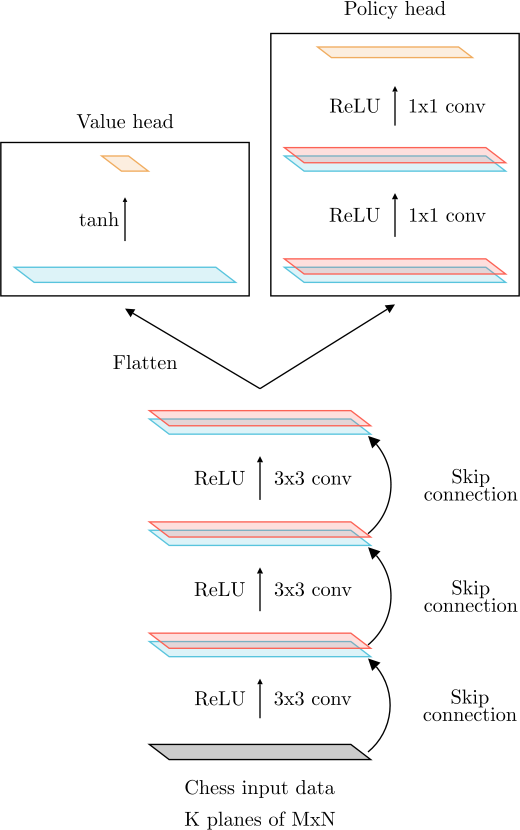}
    \caption{\label{fig:architecture} A block-diagram showing the architecture of the 6x6-ResNet model, as described in the text. The CNN-architecture for the 4x5-model has the same backbone, but without skip-connections and only three initial convolution layers. The 4x5 model has 32 filters per convolutional layer, and the 6x6-model has 64 filters per layer.}
\end{figure}

\begin{figure}
    \centering
    \begin{subfigure}{0.20\textwidth}
        \includegraphics[width=\textwidth]{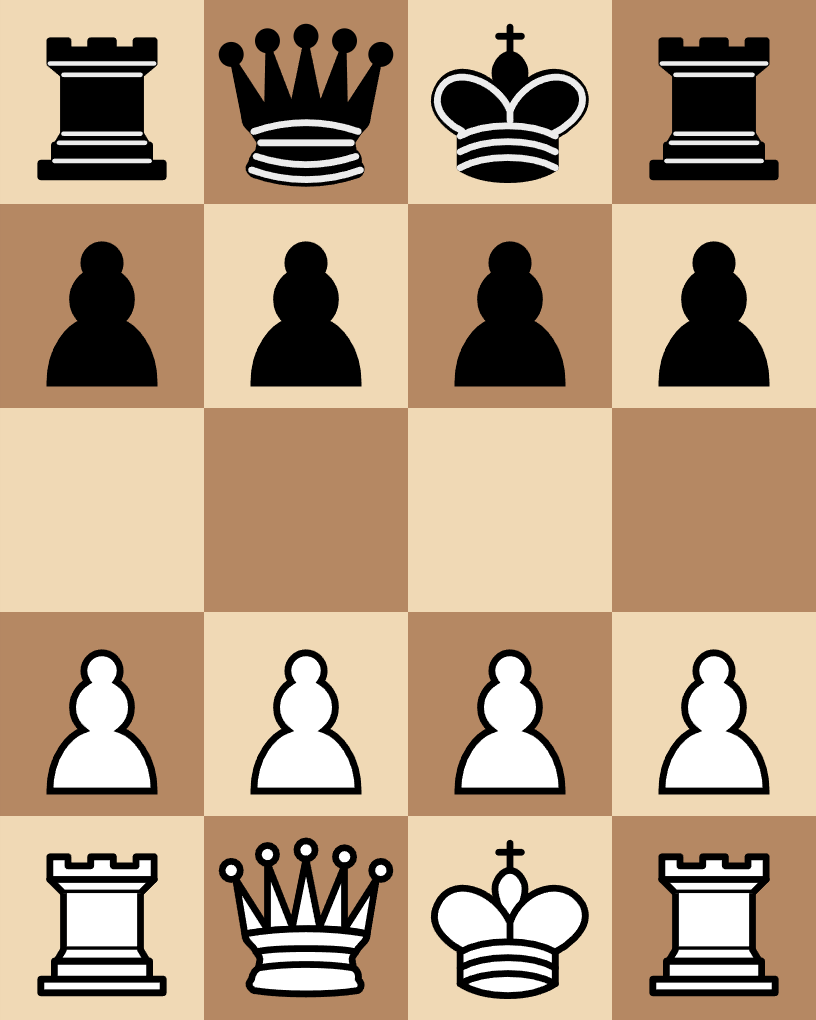}
        \caption{\label{fig:board-4x5}}
    \end{subfigure}
    \hspace{3pt}
        \begin{subfigure}{0.25\textwidth}
        \includegraphics[width=\textwidth]{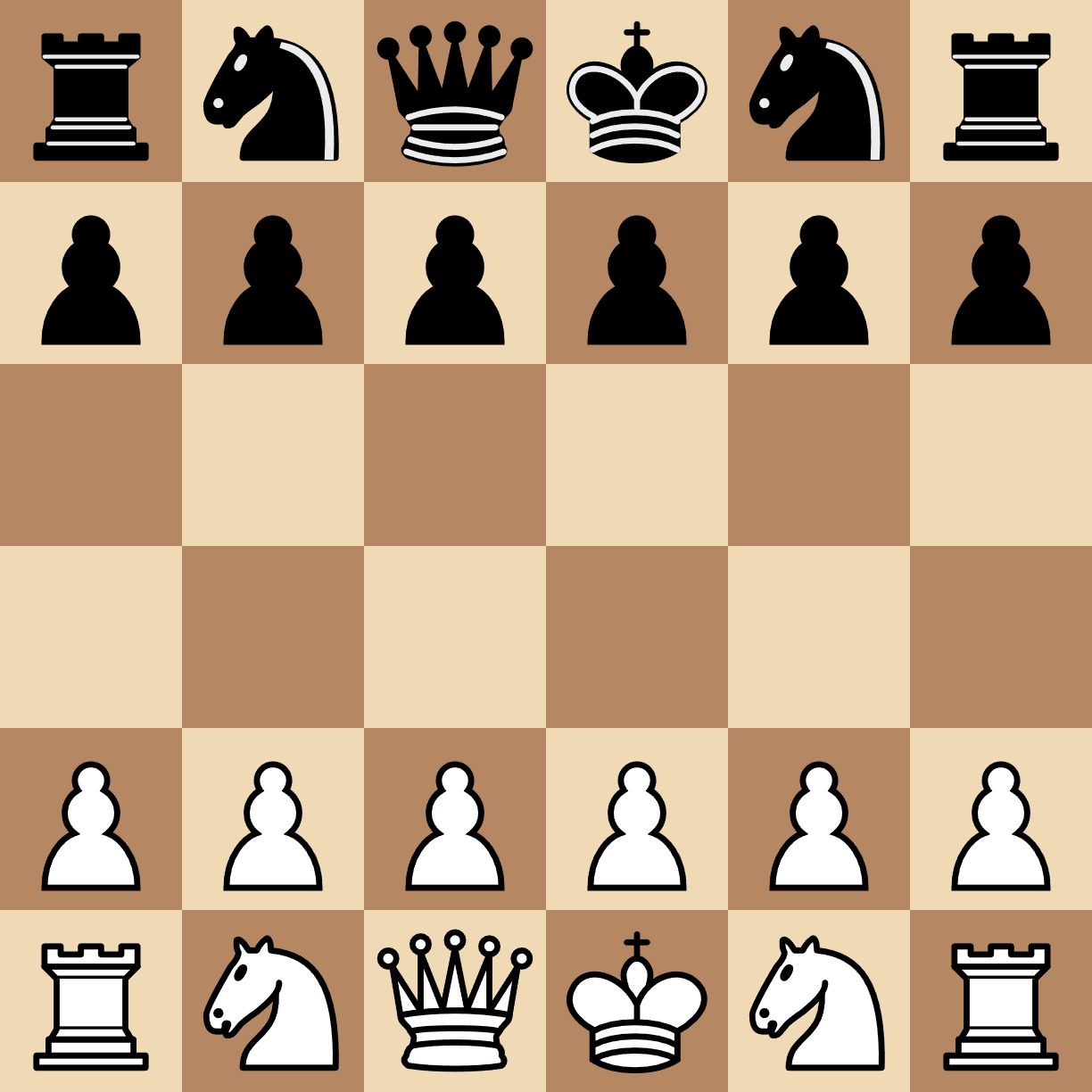}
        \caption{\label{fig:board-6x6}}
    \end{subfigure}
    \caption{The two chess-variant implementations used in this work:
    (\protect\subref{fig:board-4x5}) Silverman 4x5 and
    (\protect\subref{fig:board-6x6}) Los Alamos 6x6.
    }
    \label{fig:boards}
\end{figure}

In this work, we present two agents; one trained on Silverman 4x5 chess, seen in Fig.~\ref{fig:board-4x5},
and one trained on Los Alamos 6x6 chess, seen in Fig.~\ref{fig:board-6x6}.
Both agents are feed forward convolutional neural networks (CNNs), first described by \cite{cnn}. The 6x6 model has added residual connections, akin to the architecture of AlphaZero~\cite{alphazero_original}. This has its origins in a standard ResNet-architecture, introduced by \cite{resnet}. The main architecture for both the models is shown in Fig. \ref{fig:architecture}.

Despite the reduced board size, compared to the usual 8x8 chess, both the 4x5 and the 6x6 chess variants are quite challenging to master. Our 4x5 agent is trained for 100 iterations, each iteration involving 250 games played. The 6x6 agent is trained for 400 iterations, each iteration involving 500 training games played. Both require tuning of the models' and the training loops' hyperparameters.\footnote{The full models and code used to produce them are available at \url{https://github.com/patrik-ha/explainable-minichess}}
By building the full search tree, we have found that optimal play results in a draw for the Silverman 4x5 variants.
Our 4x5-agent reaches this strategy of play, producing draws in almost all simulation games, which we -- together with the stagnating validation loss -- interpret as having identified and settled on a stable and ideal playing style.
In the following section, we provide further details about the training environment.
\subsection{Environment}
We provide a chess environment for reinforcement learning, with user-defined board-size and starting position, and a simple interface for extracting information about the internal state. The environment also allows customising castling-rules, and creates a representation of each position conforming to the input standard used by AlphaZero. This is highly relevant when used in conjunction with an RL training-loop, and also includes creating a one-to-one mapping for moves between the state-space of the environment, and the state space used for representation in the neural net. 
The environment is designed for easy extendability. Although it currently only implements standard chess pieces, it can easily be extended with custom move sets and rules. This is particularly relevant for smaller chess variants, where additional artificial rules may be added in order to increase complexity without increasing the dimensionality. 

The chess environment is written primarily in Python. We use a standard bitboard representation first proposed by \cite{bitboardorigin}, enhanced with ``magic-move bitboard generation'' introduced in \cite{kannan2007magic}, providing a significant speedup in move generation. These are both standard techniques for building compact and efficient chess simulators.
Python was chosen due to its ease of implementation, and convenience for the wider machine learning community. Although compiled languages, such as C++, are often faster, having access to the backbone of the environment allows direct access to computations done by the main game-play loop, which will be required later, see Sec. \ref{sec:concepts}. Almost all operations performed during game-play are bit-operations, owing mainly to the bitboard representation of the game. These low-level operations are done on NumPy-integers \citep{numpy}, since most bit-operations on these have bindings that are pre-compiled in C. This alleviates most of the performance loss from using Python. Some parts of the environment are compiled using Numba \citep{numba2015}, effectively providing a JIT-compiled version binding of the Python-code in C.

The environment is also coupled with a standard variant of Monte Carlo Tree Search (MCTS), first described in \cite{mcts}. The implementation utilises simple multi-threading, with minimal communication between the threads. In particular, a tree-paralellized version of MCTS is used, described in \cite{parallelmcts}. Additionally, our implementation exploits the fact that since our neural network is only updated after a given number of self-play iterations, subsequent simulation games between these updates are independent. TensorFlow Lite \citep{tensorflow2015-whitepaper} is utilised to provide fast neural network guidance without having to rely on batching predictions for the GPU. This is doubly important since our implementation requires a separate instance of the predictive model per thread.

This environment is used to train the two agents described in Sec.~\ref{sec:agents}.

\subsection{Concept Detection}\label{sec:concepts}
We use the concept detection method used in \cite{alphazero_concepts_2021}, based on the idea introduced in \cite{tcav_2018}. Briefly described, concepts are detected by learning a set of logistic probes for a data set representing the concept of interest, using the activation outputs for each intermediary layer in the neural network. In the binary case, for a given concept \(C\), an intermediary layer of size \(m\) in the neural network, a given list of inputs \({I_0},{I_1}...{I_N}\), the corresponding activation outputs \({O_0},{O_1}...{O_N}\), and binary labels \({P_i} \in \left\{ {0,1} \right\}\) for each activation output, the aim is to find the best-fitting logistic regressor with weights \(\mathbf{w}\) and bias \(\mathbf{b}\) so that 
\begin{equation}
    \left\| {\sigma \left( {\mathbf{w} \cdot {O_i} + \mathbf{b}} \right) - {P_i}} \right\|_2^2
\end{equation}
is minimized for all \({P_i}\) and \({O_i}\). Here, \(\sigma \) is the sigmoid function. To ensure that the concepts found are actually compactly represented, an $L1$-penalty weighted by \(\lambda \) is added to the weights, which in turn gives the final minimization objective:
\begin{equation}
    \left\| {\sigma \left( {\mathbf{w} \cdot {O_i} + \mathbf{b}} \right) - {P_i}} \right\|_2^2 + \lambda {\left\| \mathbf{w} \right\|_1} + \lambda \left| \mathbf{b} \right| \,.
\end{equation}

The presence of a concept in different layers of the neural network is calculated by splitting the data set representing the concept into a training set and a validation set, learning the logistic regressor on the training set and calculating its binary accuracy on the validation set.
In this case, for the logistic probe \(L\left(  \cdot  \right)\), the binary accuracy, corrected for random guessing, is 
\begin{equation}
    \frac{2}{N} \left( {\sum_i^N {{H\left( {L\left( {{O_i}} \right) - 0.5} \right) - {P_i}} } } \right) - 1 \,,
\end{equation}
where $H$ is the Heaviside step function.

\begin{table}[hb]
    \centering
    \captionsetup{width=.9\linewidth}
    \caption{\label{table:concepts}Concepts probed in the chess agents.}
    \begin{tabular}{ll}
    \toprule
        Name & Description \\
        \midrule
        \texttt{has\_mate\_threat} & Checkmate is available \\
        \texttt{in\_check} & Is in check \\
        \texttt{material\_advantage} & Has more pieces than opponent \\
        \texttt{threat\_opp\_queen} & Opponent's queen can be captured\\
        \bottomrule
    \end{tabular}
    
\end{table}

We probe the two agents for four concepts in total, listed in Table \ref{table:concepts}. The concept probing data sets are created through large amounts of self-play between all of our model checkpoints. Then, a subset containing 10\% of the positions is randomly sampled from these games, and the positions labelled according to the concept they represent. This processed is repeated until enough positions are gathered to form a balanced data set\footnote{We perform a sanity check of our concept probing approach applying it to a data set with random concept labels, for which it predicts approximately $0$ in all layers.}.
This differs from the approach in \cite{alphazero_concepts_2021}, where concept data sets are obtained from a database of expert-level games.
We choose our approach for two reasons: Primarily, there is no database of recorded games for smaller chess variants. Secondly, we wish to be able to generate balanced data sets of arbitrary sizes, which is not possible when sampling from a fixed-size data set. 
To avoid any bias in the concept data set in favour of the models being tested, we sample from a large variety of models, in addition to adding noise to the move selection process. This noise also ensures that each model is able to produce a large amount of potential games.

Each of our concept data sets consists of \(2.5 \cdot {10^5}\) positive and \(2.5 \cdot {10^5}\) negative samples. The validation ratio is \(0.2\), and the $L1$ weighting is \(\lambda  = 0.01\).

\section{Results}\label{sec:results}
\begin{figure}
    \centering
    \begin{subfigure}{0.24\textwidth}
        \includegraphics[width=\textwidth]{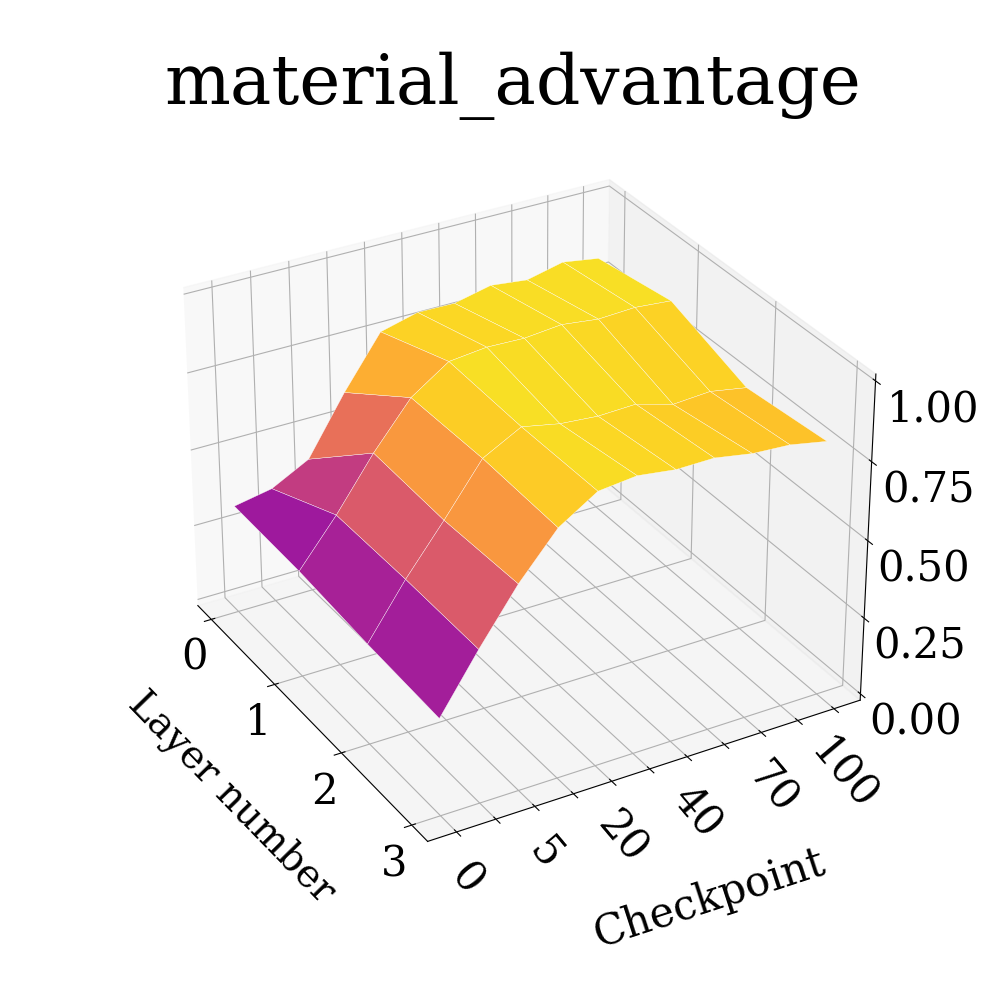}
        \caption{}
        \label{fig:concepts4x5-material}
    \end{subfigure}
    \begin{subfigure}{0.24\textwidth}
        \includegraphics[width=\textwidth]{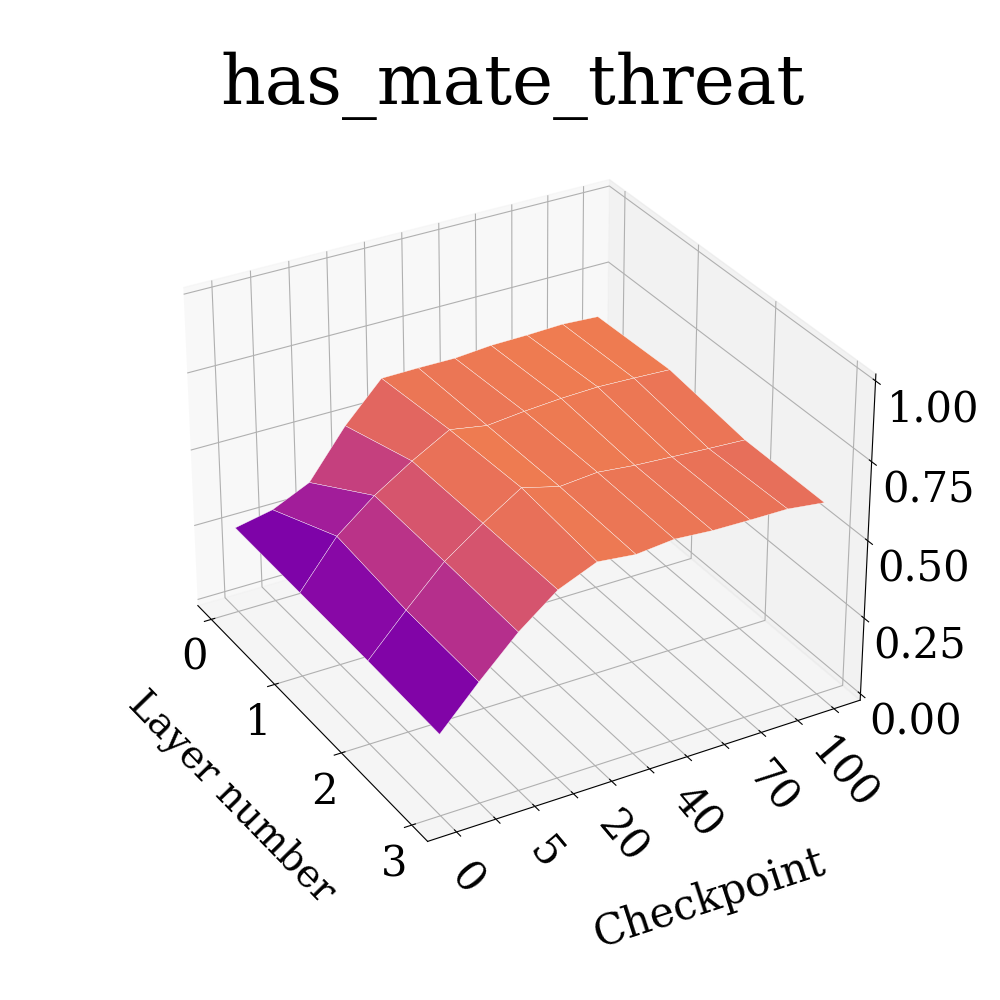}
        \caption{}
        \label{fig:concepts4x5-mate-threat}
    \end{subfigure}
        \begin{subfigure}{0.24\textwidth}
        \includegraphics[width=\textwidth]{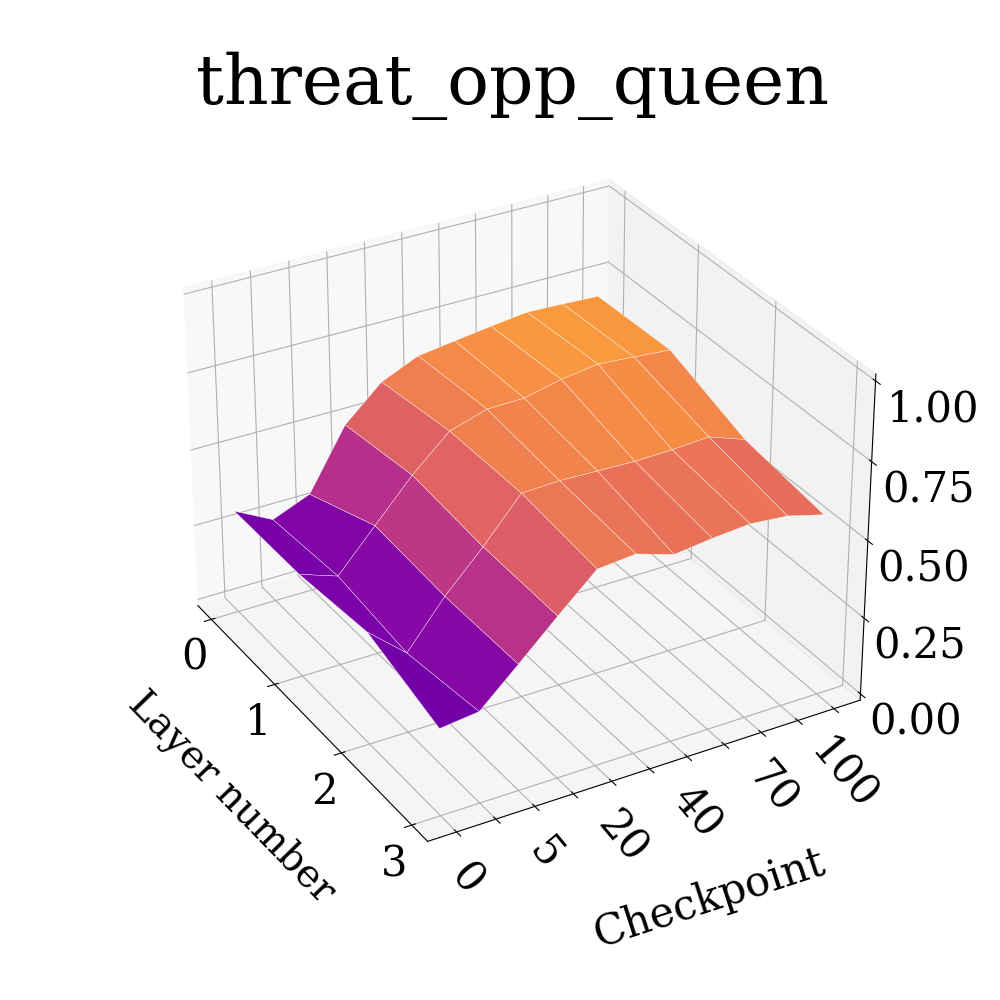}
        \caption{}
        \label{fig:concepts4x5-queen}
    \end{subfigure}
    \begin{subfigure}{0.24\textwidth}
        \includegraphics[width=\textwidth]{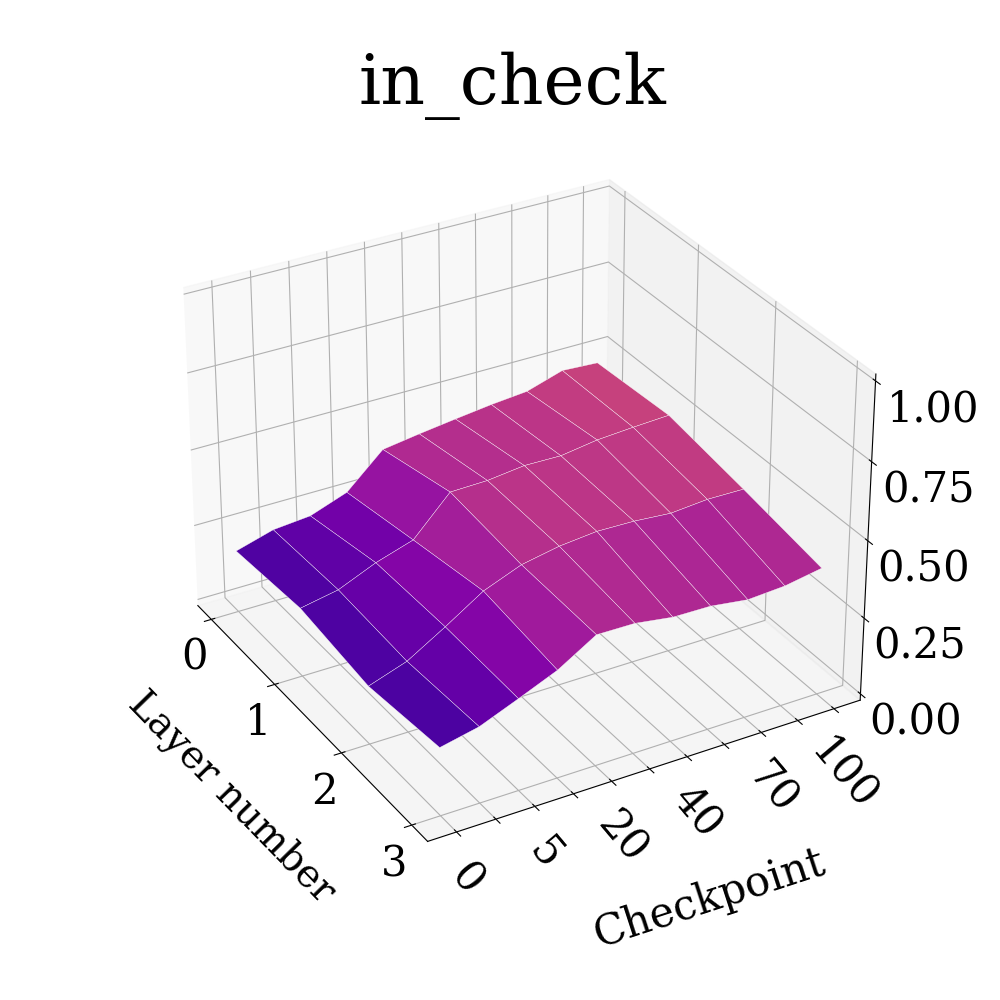}
        \caption{}
        \label{fig:concepts4x5-in-check}
    \end{subfigure}
    \caption{Concepts modelled by the 4x5 chess board agent, showing development of the model's ability to detect
    (\protect\subref{fig:concepts4x5-material}) whether the player to move has a significant material advantage, 
    (\protect\subref{fig:concepts4x5-mate-threat}) whether the opponent could mate it in a single move,
    (\protect\subref{fig:concepts4x5-queen}) whether the opponent's queen is threatened,
    and
    (\protect\subref{fig:concepts4x5-in-check}) whether it is in check,
    developing in the course of the training checkpoints.
    }
    \label{fig:concepts4x5}
\end{figure}

\begin{figure}
    \centering
    \begin{subfigure}{0.24\textwidth}
        \includegraphics[width=\textwidth]{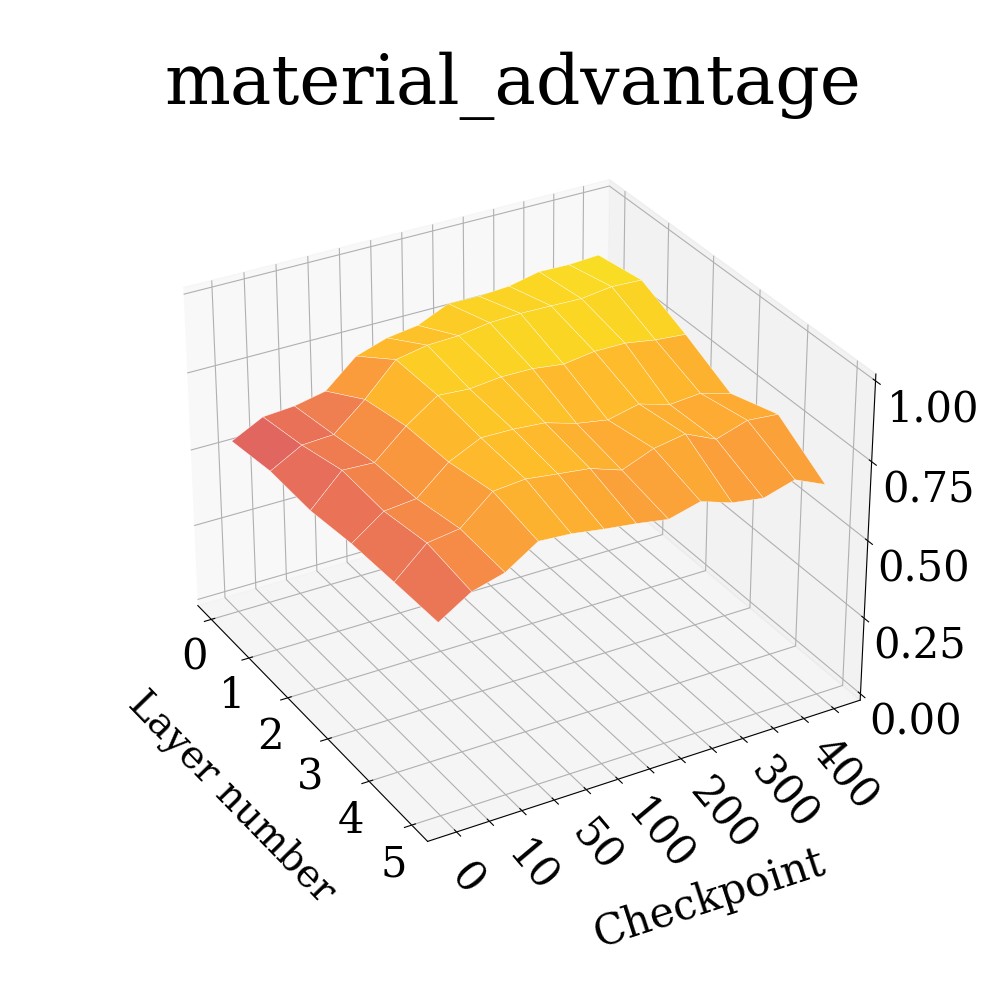}
        \caption{}
        \label{fig:concepts6x6-material}
    \end{subfigure}
    \begin{subfigure}{0.24\textwidth}
        \includegraphics[width=\textwidth]{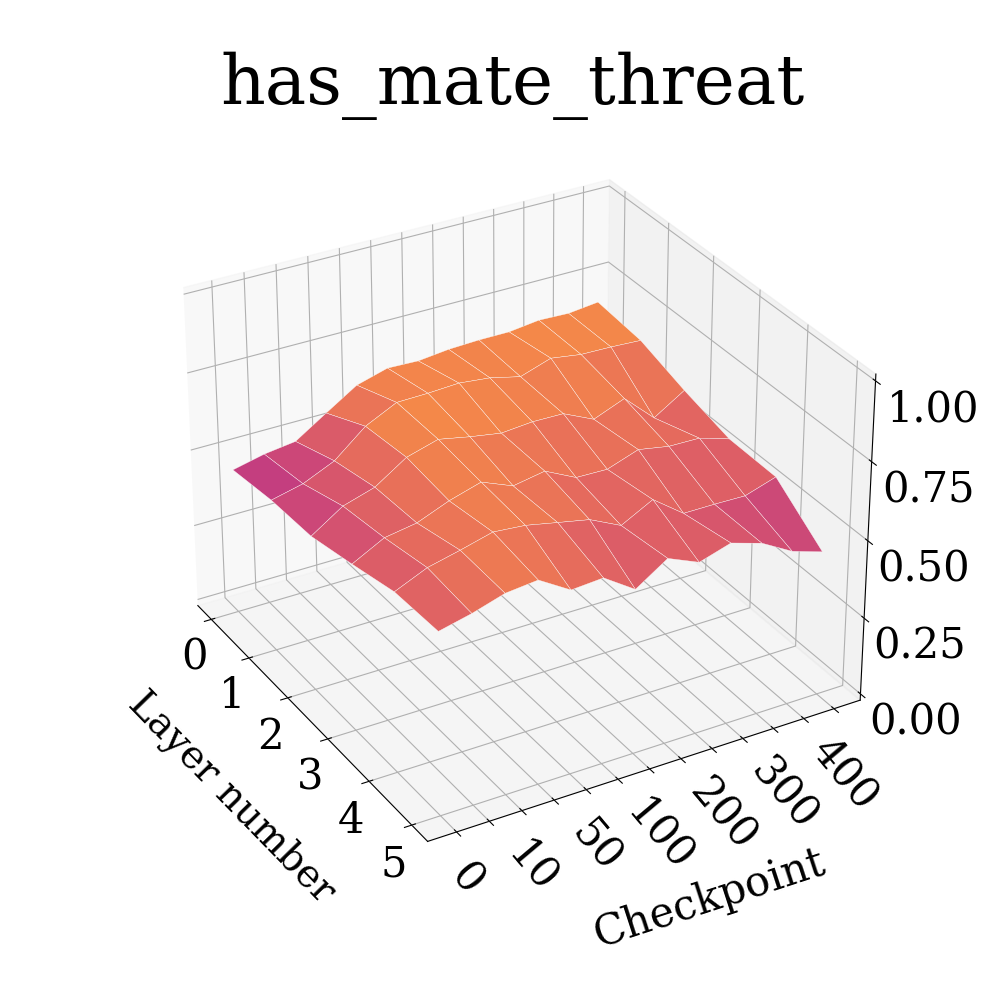}
        \caption{}
        \label{fig:concepts6x6-mate-threat}
    \end{subfigure}
        \begin{subfigure}{0.24\textwidth}
        \includegraphics[width=\textwidth]{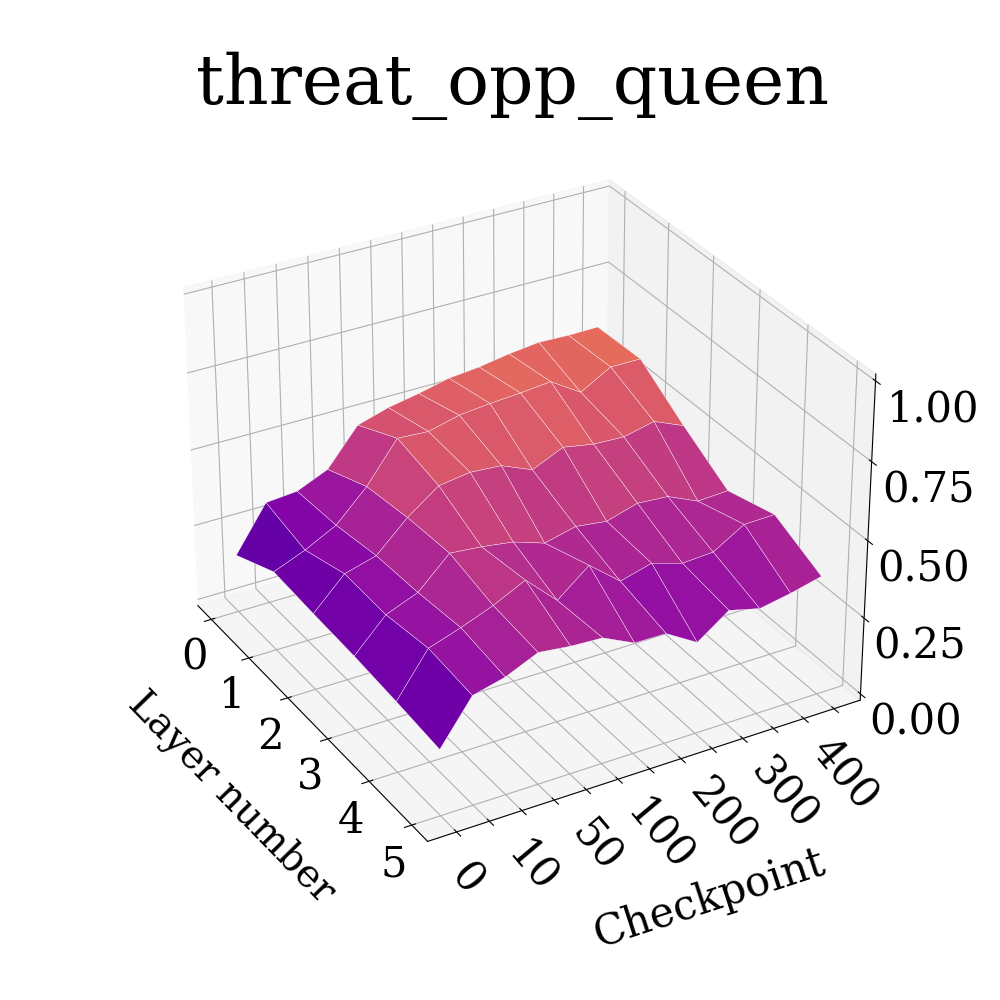}
        \caption{}
        \label{fig:concepts6x6-queen}
    \end{subfigure}
    \begin{subfigure}{0.24\textwidth}
        \includegraphics[width=\textwidth]{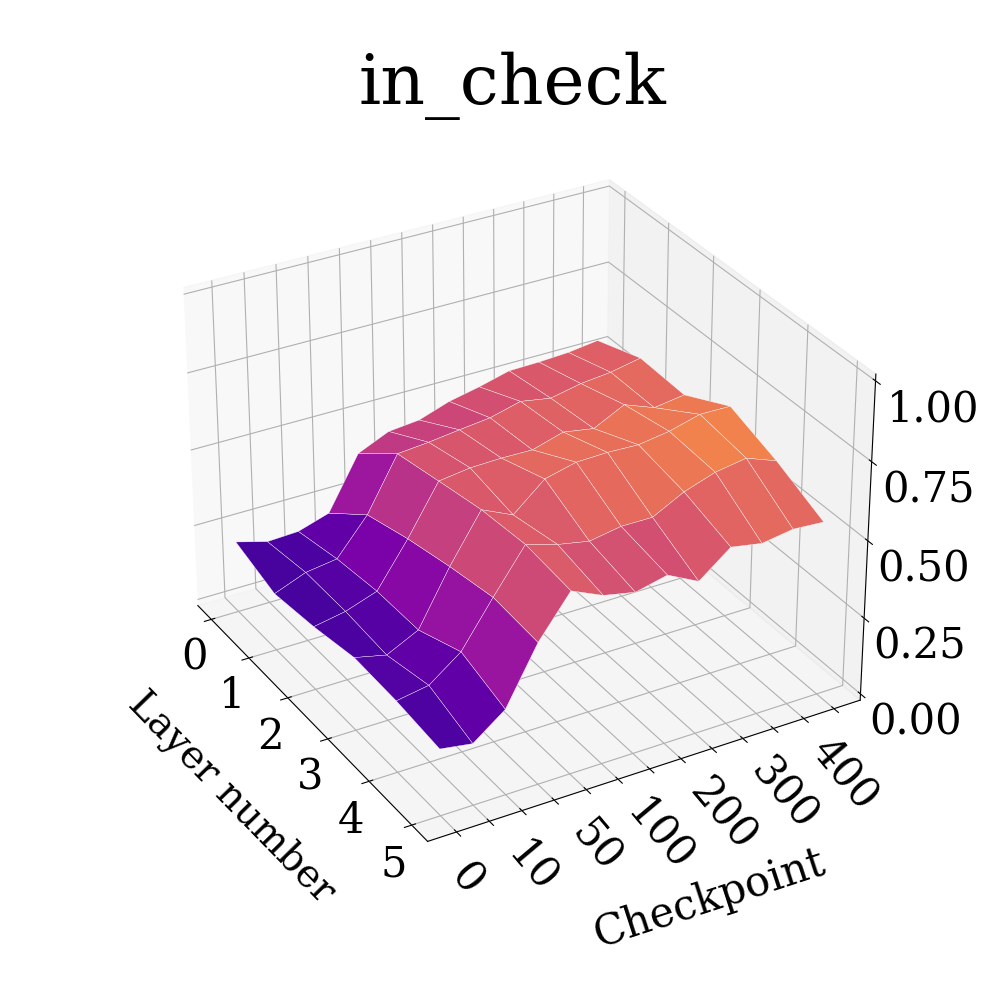}
        \caption{}
        \label{fig:concepts6x6-in-check}
    \end{subfigure}
    \caption{Concepts modelled by the 6x6 chess board agent, showing development of the model's ability to detect
    (\protect\subref{fig:concepts6x6-material}) whether the player to move has a significant material advantage, 
    (\protect\subref{fig:concepts6x6-mate-threat}) whether the opponent could mate it in a single move,
    (\protect\subref{fig:concepts6x6-queen}) whether the opponent's queen is threatened,
    and
    (\protect\subref{fig:concepts6x6-in-check}) whether it is in check,
    developing in the course of the training checkpoints.
    }
    \label{fig:concepts6x6}
\end{figure}

The probed concepts for the 4x5 and 6x6 agents are shown in Figs.~\ref{fig:concepts4x5} and~\ref{fig:concepts6x6}. Here, we see concepts evolving over the course of training. 
In Fig.~\ref{fig:concepts4x5-material}, we see that the 4x5-agent develops a representation for detecting whether it has a material advantage, meaning the ability to count the material value of the opponent's pieces compared to its own.
Soon after in the training process, proceeding approximately 30 iterations, the 4x5-agent learns to represent whether it is threatening the opponent's queen, see Figs.~\ref{fig:concepts4x5-queen}. 
Later, after approximately 50 iterations, the agent learns to represent whether it has a potential mating attack, see Figs.~\ref{fig:concepts4x5-mate-threat}. 
Surprisingly, we observe that the agent represents the state of being in check only weakly throughout the entire training process, see Fig.~\ref{fig:concepts4x5-in-check}, despite learning to play optimally.
Observe that all the probed concepts flatten out after about 100 training iterations, regardless of whether training continues. This indicates that this agent is highly unlikely to allocate more resources to represent these concepts provided more experience.

For the 6x6-agent, we see many of the same trends as for the 4x5 agent, compare Figs.~\ref{fig:concepts6x6-material}, \ref{fig:concepts6x6-mate-threat} and \ref{fig:concepts6x6-queen}. 
A main difference is that \texttt{material\_advantage} as well as \texttt{has\_mate\_threat} are detectable already at the beginning of the training loop. This is likely due to the residual connections combined with the increased model size of the 6x6-agent, providing it with the capacity to represent the concepts, even in an untrained state. For models with skip connections, it is in general more likely that information from the data is maintained throughout the model, albeit unintentionally.
We also see that this agent learns to detect whether it is in check, see Fig.~\ref{fig:concepts6x6-in-check}, in contrast to the 4x5-agent. Both agents weakly represent this concept during the early stages of training, but only the 6x6-agent continues to develop this concept throughout the entire training process.

\section{Analysis}\label{sec:analysis}
In an RL context, most of the early learning takes place because the agent accidentally succeeds, i.e.\ delivers checkmate. As a consequence, concepts connected to actions that are statistically more likely to win the game for a given agent, appear during the earliest stages of training.
For the 4x5-agent, we observe that it quickly creates and bolsters an internal representation of material imbalance, see Fig.~\ref{fig:concepts4x5-mate-threat}. The reason is most likely that this a simple yet highly predictive proxy for which player will win. For very weak agents, doing mostly random action selection strategies, it is highly beneficial to have more pieces than the opponent, as it implies controlling more squares than the opponent, which in turn gives a higher chance of winning. This is also reflected in the appearance and development of the ability to detect threats to the opponent's queen. On average, the queen controls the largest number of squares per piece, and therefore serves as a simple proxy for ``the ability to deliver checkmate'' when playing in a non-informed way. The same tendency was observed in \cite{alphazero_concepts_2021}.

We also observe the emergence of a canonical concept for the 4x5-agent, representing whether the player has a guaranteed win, shown in Fig.~\ref{fig:concepts4x5-mate-threat}. However, for the same agent, we do not see a strong emergence of the ability to detect whether the player to move is in check, despite the model seemingly understanding the concepts of threats and attacks (see Figs.~\ref{fig:concepts4x5-mate-threat} and~\ref{fig:concepts4x5-queen}). We expected that being in check would be an important concept to represent, as being in check significantly limits what moves are available for play. For the model to be able to predict legal moves in this situation, it would therefore be reasonable to assume that it needs to detect if it is in check. The reason for the weak representation of this concept is most likely that the ratio of legal to illegal moves is significantly smaller on this board than on 6x6, or even 8x8-boards. This would also explain why this concept is more strongly represented in our 6x6-agent model (see Fig.~\ref{fig:concepts6x6-in-check}), and in \cite{alphazero_concepts_2021}.

In the case of the 6x6-agent, we observe a localised emergence of concepts. In Figs.~\ref{fig:concepts6x6-queen} and~\ref{fig:concepts6x6-in-check}, either concept emerges primarily in the first and last layers. This is due to the architecture of the 6x6-agent's model, and agrees with the interpretation presented in \cite{linearclassifierprobes} as well as the observations in \cite{alphazero_concepts_2021}. For abstract concepts, like \texttt{threat\_opp\_queen}, it makes sense that these are represented early in the model. On the other hand, it makes sense that concepts that have immediate consequences for which moves are available, like \texttt{in\_check}, are represented close to the relevant output-heads.

We also observe that there is a difference between the two agents regarding \emph{when} during training each concept emerges. We also see that the concepts represented in the smaller agent reach a plateau. This is in contrast to the larger agent, which continues developing concepts even after finding a winning playing style. This is as expected, since a larger dimensionality implies that more examples and thus training time are needed to experience a sufficient number of situations in which the different concepts are relevant.

\FloatBarrier
\section{Discussion}\label{sec:discussion}
We have demonstrated that the concept probing method is capable of detecting whether deep neural network models contain a compact, linear representation of a given concept. 
This allows probing neural network based agents for domain knowledge without having to include it in the training loop. 
This method is therefore highly relevant for investigating, evaluating and explaining neural network based autonomous agents, as it reveals whether these have internalised domain knowledge as expected. In the following, we discuss challenges and limitations of the method.

While this was not a problem in our case, generating representative concept data sets might prove difficult. Ideally, the concept must be defined programmatically, so that concept examples can be distilled from some large sample-repository. This might be challenging, especially for large and complex problems. Also, it requires an efficient simulation-environment. In cases where the environment is a simulation of a real-world application, the simulation must also be able to generate concept samples that are representative of what the agent would encounter in its ``true environment''. Otherwise, concept detection might be hindered by subtleties of synthetic data.

Exploration is a fundamental aspect of the RL paradigm, opening up the possibility that the model has observed states that, from a domain expert's viewpoint, are unlikely to occur. This is important because it makes it more likely that the model is allowed to generalise its concepts, as opposed to if it was trained, e.g., in a supervised fashion on a curated data set. 

In our implementation, we limit ourselves to probing for \emph{linearly} represented concepts, although there are no guarantees that the network represents its knowledge in a linearly separable manner. However, as observed in \cite{linearclassifierprobes}, it is expected that most neural networks with sufficient depth have will end up representing emergent concepts as simply as possible, i.e.\ linearly. 

Finally, we wish to highlight is the difference between being able to detect concepts, and knowing how the represented concepts are utilised within the given model. In our case, we see that both models develop some representation of \texttt{material\_advantage}. For a human, this particular concept is usually used as a heuristic for the result of the game, meaning that the player holding fewer pieces is more likely to lose. However, we cannot guarantee that our model has made the same connection, although this aligns most readily with the relevant domain knowledge. 
\section{Conclusion}
In this work, we have demonstrated that concept probing can be used to gain insight into what concepts a neural network model learns. We have also demonstrated the feasibility of using these methods to reason whether the model's internal representations align with existing domain knowledge. This is highly relevant for all control contexts in which an autonomous agent has learned to navigate in an environment about which humans possess domain knowledge. This implies state-of-the-art RL methods can be used to solve control tasks, without losing the ability to investigate how the learned agent understands its environment. This is particularly useful for problems where modelling the environment is challenging. Given that the agent performs its task well, this can also be used to provide new understanding of the environment in question.

In the context of neural network based agents, the concept detection method is both architecture and environment agnostic. As long as it is possible to define a concept function, the method can be applied to any layered model structure. This implies easy applicability in control contexts, without having to apply constraints on the environment or training loop. While we demonstrate the applicability of the concept detection methods to agents trained in our environment, an important next step will be applying the method to classical control problems, comparing the extracted concepts to domain knowledge. 

\begin{ack}
We thank Vilde B.\ Gj{\ae}rum for valuable discussions and proof reading.
\end{ack}

\bibliography{ifacconf}


\end{document}